\documentclass{article}
\usepackage{arxiv}

\usepackage{graphicx}
\usepackage{color}
\usepackage{xurl}
\usepackage{bm}
\usepackage[psamsfonts]{amssymb}
\usepackage{algorithm}
\usepackage[noend]{algpseudocode}
\usepackage{algorithmicx}
\usepackage{comment}
\usepackage{amsmath}
\usepackage{multicol}
\usepackage{multirow}

\usepackage{dcolumn}
\usepackage{lipsum}

\usepackage{fixltx2e}

\usepackage[hang,small,bf]{caption}
\usepackage[subrefformat=parens]{subcaption}
\algdef{SE}[SUBALG]{Indent}{EndIndent}{}{\algorithmicend\ }%
\algtext*{Indent}
\algtext*{EndIndent}

\title{An FPGA-Based Accelerator for Graph Embedding using Sequential Training Algorithm}
\author{
  Kazuki Sunaga \\
  Keio University \\
  3-14-1 Hiyoshi, Kohoku-ku, Yokohama, Japan \\
  \texttt{sunaga@arc.ics.keio.ac.jp} \\
  \And
  Keisuke Sugiura \\
  Keio University \\
  3-14-1 Hiyoshi, Kohoku-ku, Yokohama, Japan \\
  \texttt{sugiura@arc.ics.keio.ac.jp} \\
  \And
  Hiroki Matsutani \\
  Keio University \\
  3-14-1 Hiyoshi, Kohoku-ku, Yokohama, Japan \\
  \texttt{matutani@arc.ics.keio.ac.jp}
}
\begin{document}
\maketitle

\begin{abstract}
A graph embedding is an emerging approach that can represent a graph structure with a fixed-length low-dimensional vector. 
node2vec is a well-known algorithm to obtain such a graph embedding by sampling neighboring nodes on a given graph with a random walk technique. 
However, the original node2vec algorithm typically relies on a batch training of graph structures; thus, it is not suited for applications in which the graph structure changes after the deployment. 
In this paper, we focus on node2vec applications for IoT (Internet of Things) environments. 
To handle the changes of graph structures after the IoT devices have been deployed in edge environments, in this paper we propose to combine an online sequential training algorithm with node2vec. 
The proposed sequentially-trainable model is implemented on an FPGA (Field-Programmable Gate Array) 
device to demonstrate the benefits of our approach. 
The proposed FPGA implementation achieves up to 205.25 times speedup compared to the original model on ARM Cortex-A53 CPU.
Evaluation results using dynamic graphs show that although the accuracy is decreased in the original model, the proposed sequential model can obtain better graph embedding that achieves a higher accuracy even when the graph structure is changed.

    
\end{abstract}




\section{Introduction} \label{sec:intro}
Graph structures in which nodes are connected by edges can be seen everywhere in our life. 
For example, friendships of users in social networking services, relationships between users and purchased items in e-commerce sites, and paper citation relationships 
can be represented by such graph structures. 
Thus, there are high demands for applications that can extract, analyze, and utilize information from these graph structures. 
Although a graph structure can be represented by an adjacency matrix, the adjacency matrix cannot be directly used in statistical or machine learning based methods especially when the graph structure becomes large and sparse. 
To overcome this issue, a graph embedding is an emerging representation which can be directly used with statistical or machine learning 
methods. 

Using the graph embedding, graph structures can be represented with fixed-length low-dimensional vectors.
node2vec~\cite{node2vec} is a well-known algorithm to obtain the graph embedding by sampling neighboring node information on a given graph with a random walk technique. 
However, the original node2vec algorithm typically relies on a batch training, not online sequential training; thus, it is not suited for applications where the graph structure changes after the deployment. 
In this paper, we assume node2vec applications for IoT environments. 
To handle the changes of graph structures after the IoT devices are deployed in edge environments, in this paper we propose to combine an online sequential training algorithm with node2vec.
Since the low-cost and low-power execution is required for such IoT applications, the proposed sequential model is implemented on an FPGA device in order to significantly shorten the training time at the deployed environment. 

The rest of this paper is organized as follows. 
Section~\ref{sec:related} introduces related works, and Section~\ref{sec:design} proposes a sequentially-trainable graph embedding model and its FPGA-based accelerator. 
Section~\ref{sec:eval} evaluates the model and the accelerator in terms of the execution time, accuracy, model size, and FPGA resource utilization. 
Section~\ref{sec:conc} summarizes our contributions.


\begin{figure}
	\centering
	\includegraphics[keepaspectratio, scale=0.39]{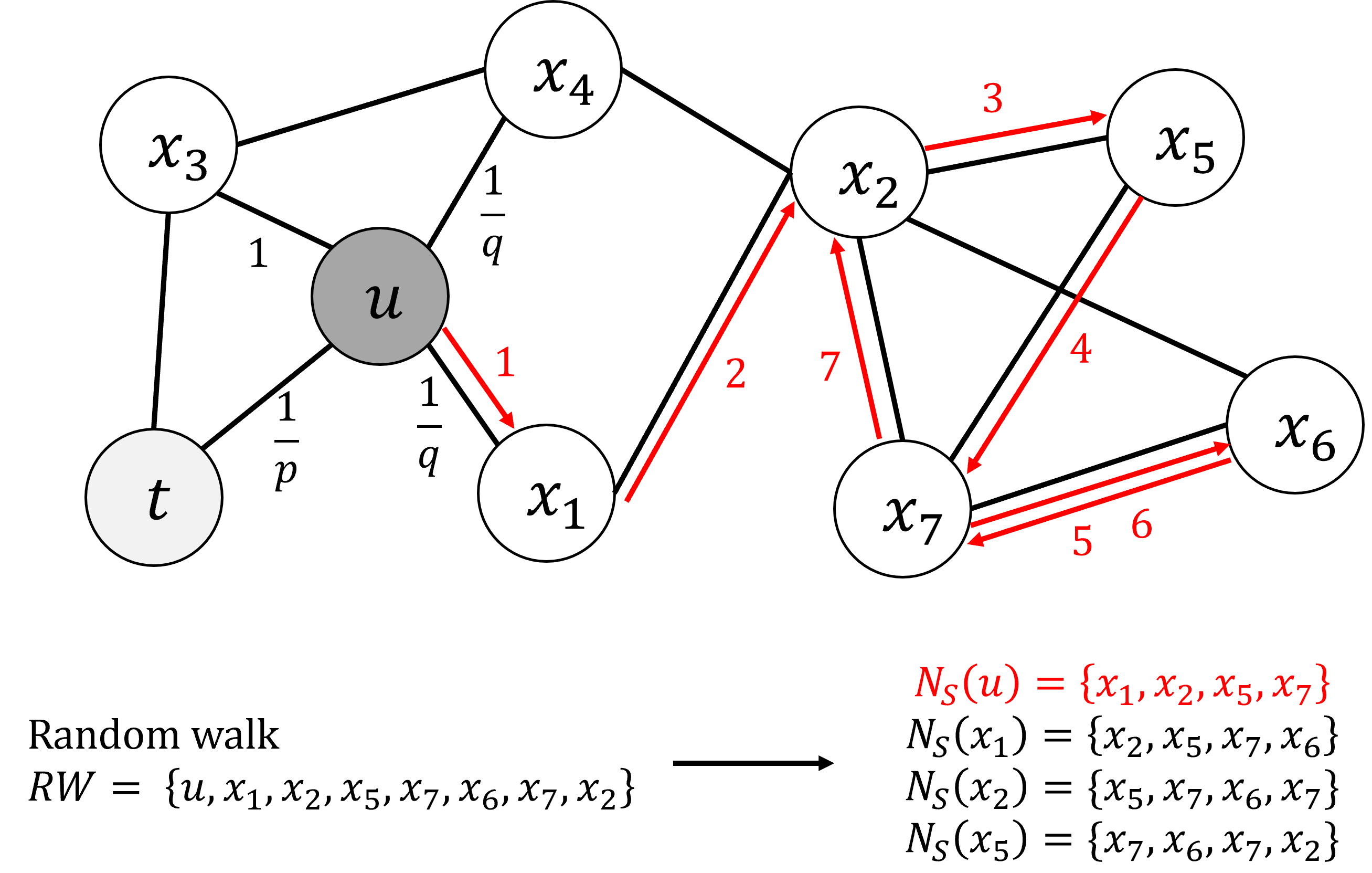}
	\caption{An example of random walk in node2vec}
	\label{fig:node2vec}
\end{figure}

\section{Related Work} \label{sec:related}
\subsection{node2vec Algorithm} \label{subsec:node2vec}
node2vec~\cite{node2vec} is a well-known algorithm to obtain a graph embedding. As shown in Figure~\ref{fig:node2vec}, a random walk is performed from a selected start node (e.g., node-$t$) in the graph in order to collect neighboring nodes information. 
Assuming that a transition has been performed from node-$t$ to node-$u$, Figure~\ref{fig:node2vec} illustrates the next transition probabilities from node-$u$ to one of its adjacent nodes. 
The transition probabilities to the adjacent nodes are determined as follows.
\begin{equation}
    P(c_i = x | c_{i-1} = u) = \left\{
	\begin{alignedat}{2}
		\frac{\alpha_{pq}(t, x) w_{ux}}{Z} \quad \text{if} \quad ((u,x) \in E) \\
		0, \qquad \quad \quad \text{if} \quad ((u,x) \notin E)
	\end{alignedat}
	\right.
\end{equation}
where $c_i$ represents the $i$-th node during a random walk. 
$(u, x)$ represents an edge between node-$u$ and node-$x$, and $w_{ux}$ represents the weight of this edge. 
$E$ represents all the edges in a graph. 
$Z$ is a normalizing constant. 
$\alpha$ is defined using parameters $p$ and $q$ as follows.
$\alpha_{pq}(t,x) = 1/p$ if $d_{tx}=0$, 
$\alpha_{pq}(t,x) = 1$ if $d_{tx}=1$, and
$\alpha_{pq}(t,x) = 1/q$ if $d_{tx}=2$, 
where $d_{tx}$ represents a distance between the previous node (i.e., node-$t$) and the next node (i.e., node-$x$). 
That is, $d_{tx}$ is set to 0 when transitioning back to node-$t$; $d_{tx}$ is set to 1 when transitioning to an adjacent node of node-$t$; otherwise $d_{tx}$ is set to 2. 
Let $RW$ be the result of a single random walk. 
The training samples can be efficiently generated by partitioning $RW$ with a given context size (i.e., window size).
In Figure~\ref{fig:node2vec}, $N_S(u)$ is an example of the training samples. 
$N_S(u)$ represents neighboring nodes of node-$u$ obtained by a random walk started from node-$u$ based on a random walk strategy $S$. 
These training samples are trained with the skip-gram model~\cite{word2vec} illustrated in Figure~\ref{fig:models} (left).


In the skip-gram model, the numbers of input-layer and output-layer dimensions are the same as the number of nodes in the graph. 
The number of hidden-layer dimensions is corresponding to the number of the graph embedding dimensions to be trained. 
An input data to the skip-gram model is a one-hot vector, where only node-$u$ is 1 and the other nodes are 0. 
An output of the model is a vector, where each node represents a probability that this node appears as an adjacent node of node-$u$. 
From $N_S(u)$ we can obtain four different teacher labels, each of which is a one-hot vector, where one of nodes $x_1$, $x_2$, $x_5$, and $x_7$ is 1 and the other nodes are 0. 
Since there are four different teacher labels, the final loss value is calculated by summing up the four loss values computed using the same weights but different one-hot teacher labels.

\subsection{node2vec for Dynamic Network} \label{subsec:dynamic}
Dealing with dynamic graphs is one of important challenges in graph-based learning, and there are many approaches to learn graph embedding for dynamic graphs. Continuous-Time Dynamic Network (CTDN)~\cite{ctdn} addresses dynamic graphs by adding temporal information to the graph and imposing temporal constraints during node2vec random walks. Specifically, CTDN prohibits a movement in the time decreasing direction during the random walk, thus preserving a temporal consistency.
In \cite{snapshot}, graph snapshots are used as temporal information and combined with node2vec for link prediction. Specifically, a link prediction at time $t$ utilizes the temporal information until time $t-1$.
In dynnode2vec~\cite{dynnode2vec}, graph embedding at time $t$ is trained by using graph embedding at time $t-1$ as initial values. Although there are many prior works that learn graph embedding on such dynamic graphs, most of them rely on a skip-gram based model and a conventional batch training with backpropagation algorithm. dynnode2vec also sequentially learns graph embedding with this approach. However, in general a sequential training using the conventional backpropagation algorithm results in the loss of previous learning results. Such a phenomenon is known as a catastrophic forgetting, which reduces accuracy.
In this paper, on the other hand, we utilize an online sequential training algorithm for graph embedding on dynamic graphs. The details are described in the next section.

\subsection{Sequential Training Algorithm} \label{subsec:oselm}
\begin{figure}
	\centering
	\includegraphics[keepaspectratio, scale=0.41]{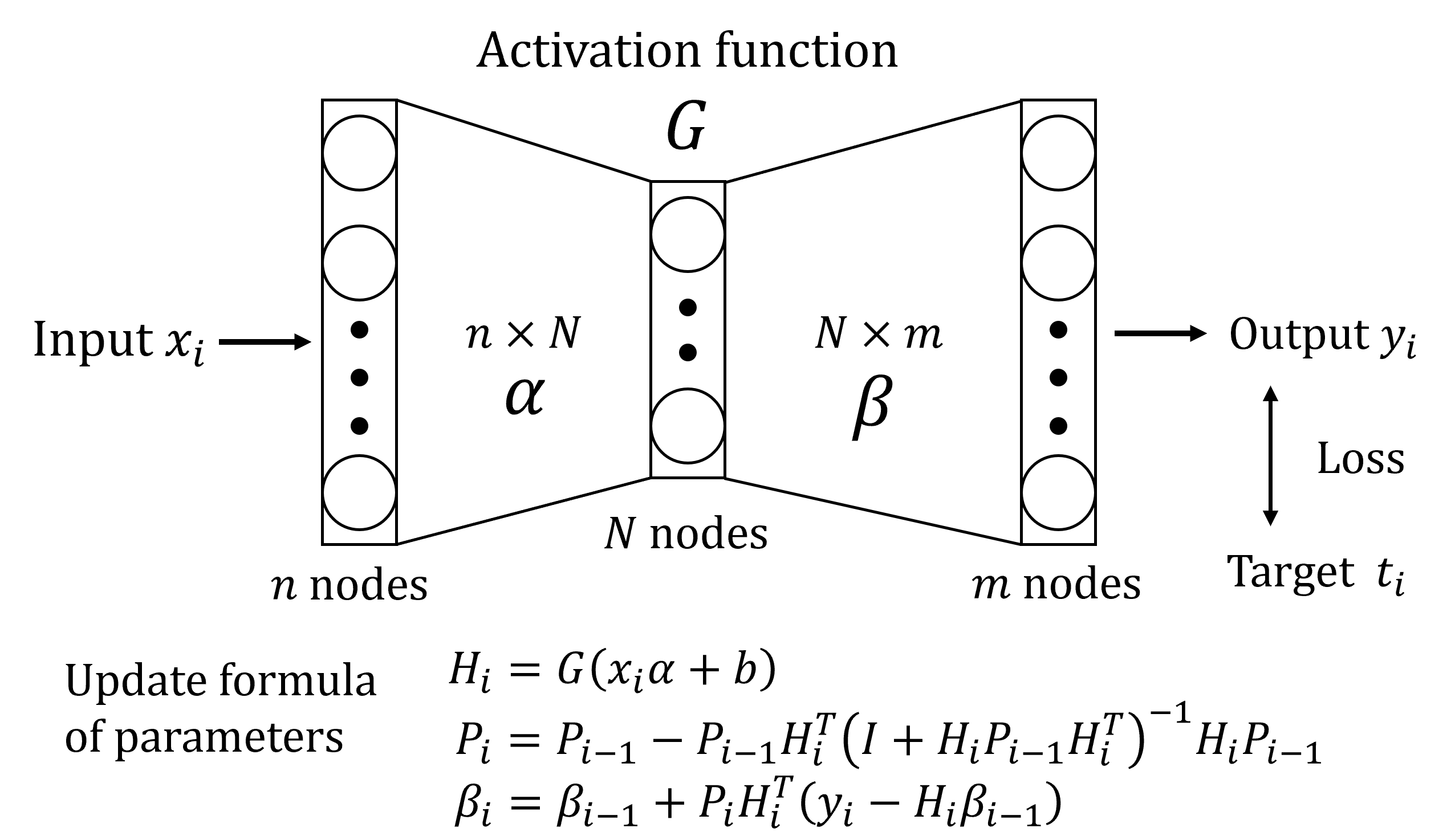}
	\caption{OS-ELM algorithm}
	\label{fig:oselm}
\end{figure}

OS-ELM (Online Sequential Extreme Learning Machine)~\cite{oselm} is an online sequential training algorithm for neural networks with a single hidden layer. 
Figure~\ref{fig:oselm} illustrates the network structure and its training algorithm.
The input-layer, hidden-layer, and output-layer dimensions are $n$, $N$, and $m$, respectively.
In the OS-ELM algorithm, the input-side weights $\bm\alpha \in \mathbb{R}^{n \times N}$ are fixed at random values at the initialization time, and only the output-side weights $\bm\beta \in \mathbb{R}^{N \times m}$ are trainable and sequentially updated. 
Assuming that the $i$-th input data is fed to the neural network, hidden-layer outputs $\bm{H_i} \in \mathbb{R}^{N}$ are generated. 
Then, new output-side weights $\bm{\beta_i}$ are calculated based on previous weights $\bm{\beta_{i-1}}$ and temporary values $\bm{P_i} \in \mathbb{R}^{N \times N}$, which are also calculated based on previous values $\bm{P_{i-1}}$ and $\bm{H_i}$. 
As shown in the equations in the figure, the OS-ELM algorithm derives an optimal $\bm\beta$ that can minimize a loss between final outputs $\bm{y_i} \in \mathbb{R}^{m}$ and teacher labels $\bm{t_i} \in \mathbb{R}^m$ analytically, where $\bm{y_i} = G(\bm{x_i \alpha + b})\bm\beta$. 
This sequential training is simple and fast since the sequential training is done in a single epoch, which is different from a conventional backpropagation based training.
An FPGA-based acceleration of OS-ELM is reported \cite{Tsukada20}.

\begin{figure}
	\centering
	\includegraphics[keepaspectratio, scale=0.29]{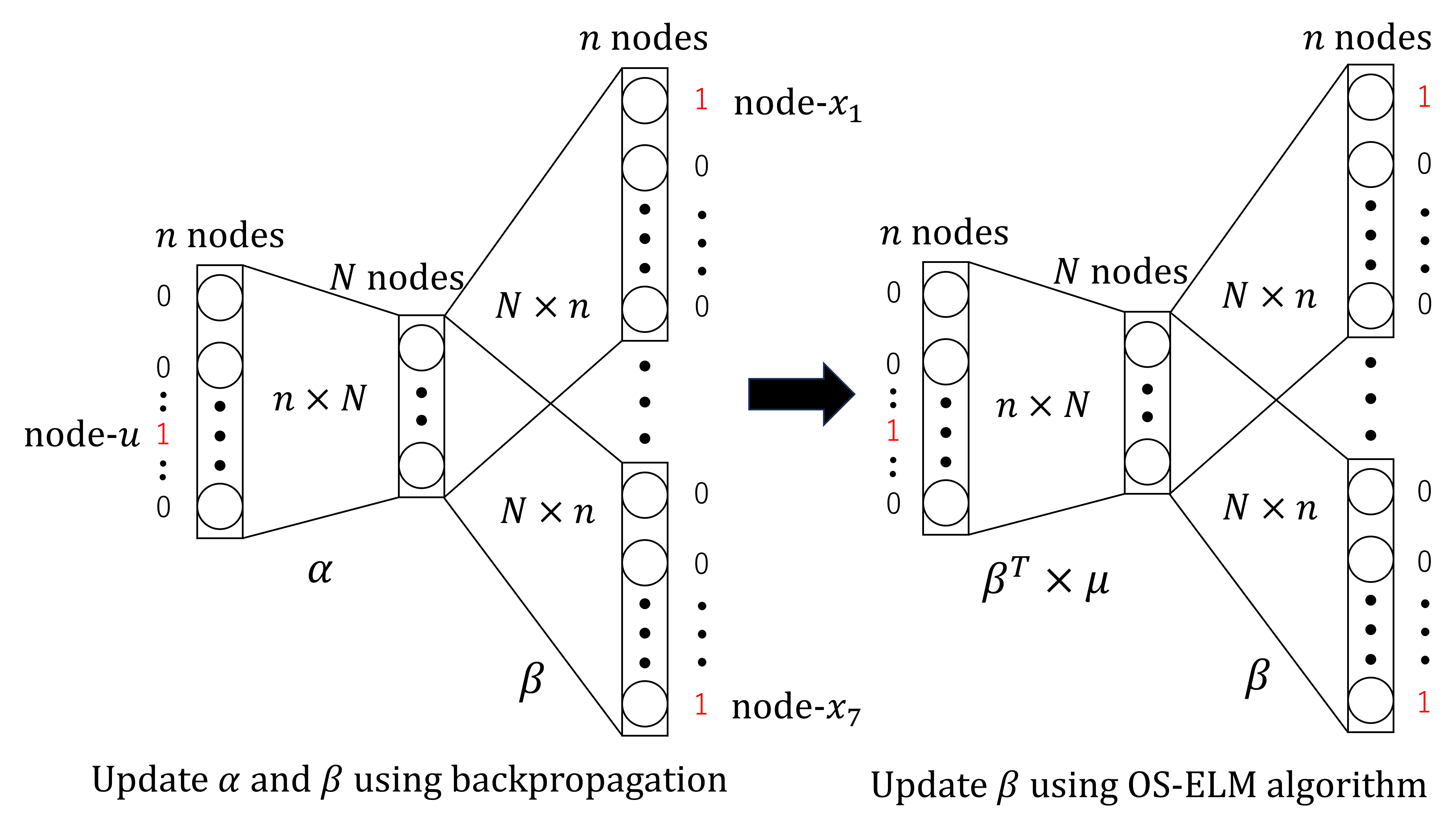}
	\caption{Original skip-gram model (left) and proposed model (right)}
	\label{fig:models}
\end{figure}

\subsection{Graph Embedding for IoT Environments}
Although graph-based learning has been actively studied in recent years, how to utilize graph embedding and graph neural networks (GNNs) for IoT applications is still an emerging research topic. For example, there are many prior works that utilize graph structures in IoT environments such as network anomaly detection, malware detection, smart transportation, and smart grids~\cite{survey1}~\cite{survey2}. It is reported that using graph structures can achieve favorable results in these tasks. On the other hand, since real-world applications typically require low latency while training large amounts of data, acceleration methods of the training have been studied so far. These training acceleration methods include algorithmic approaches and hardware based approaches, and they are classified in~\cite{accelerator}. 
In this paper, we utilize node2vec, and some prior works that utilize node2vec for IoT tasks have been reported. For example, node2vec is used to detect malware in IoT environments~\cite{malware}. It is also used to predict usage patterns in IoT environments~\cite{IoT}.

\subsection{node2vec Accelerator} \label{subsec:fpga_impl}
Hardware accelerations of graph learning methods have been studied so far. In this section, we introduce those related to node2vec.
For example, an FPGA-based acceleration of random walk for node2vec is reported~\cite{lightrw}. 
An FPGA-based acceleration of word2vec~\cite{word2vec} that uses the skip-gram model as well as node2vec is reported~\cite{word2vec-fpga}. 
Please note that the random walk is accelerated in~\cite{lightrw} while in this paper we accelerate the training algorithm of node2vec. 
Although the training algorithm of word2vec is accelerated by FPGA in~\cite{word2vec-fpga}, in this paper we newly propose a sequential training of node2vec for dynamic graphs by combining OS-ELM and the skip-gram model so that the on-device training can be efficiently implemented on FPGA devices.


\section{Sequential Graph Embedding Accelerator} \label{sec:design}

\subsection{Sequentially-Trainable Skip-Gram Model} \label{subsec:model}
Figure~\ref{fig:models} (right) illustrates the proposed OS-ELM based training model for graph embedding, and Algorithm~\ref{alg:propose} describes the training algorithm. 
Since both the skip-gram and OS-ELM assume neural networks with a single hidden layer, the OS-ELM algorithm can be theoretically applied to the skip-gram. 
In the original OS-ELM, $\bm{H_i}$ in Algorithm~\ref{alg:propose} are calculated using $\bm{x_i} \in \mathbb{R}^n$; specifically, $\bm{H_i} = G(\bm{x_i \alpha + b})$. 
Since the input vector $\bm{x_i}$ is one-hot, $\bm{H_i}$ can be  calculated as the row vector corresponding to the center node which is extracted from $\bm\alpha$ assuming that $\bm b$ is zero. 
\begin{figure}[t]
\begin{algorithm}[H]
	\caption{Proposed algorithm}
	\label{alg:propose}
	\begin{algorithmic}[1]
		\For {each context}
			\State $\bm{H_i} \leftarrow \bm\beta[$center node$] \times \mu$
			\State Compute $\bm{P_{i-1}H_i^T}$ and $\bm{H_iP_{i-1}}$
			\State Compute $\bm{P_{i-1}H_i^TH_iP_{i-1}}$ and $\bm{H_iP_{i-1}H_i^T}$
			\State $hpht\_inv \leftarrow \frac{1}{\bm{H_iP_{i-1}H_i^T}}$
			\State $\bm{P_i} \leftarrow \bm{P_{i-1}} - \bm{P_{i-1}H_i^TH_iP_{i-1}} \times hpht\_inv$
			\State Compute $\bm{P_{i}H_i^T}$
			\For {each window}
				\For {$itr=1$ to $ns+1$}
					\If {$itr = 1$}
						\State{$sample \leftarrow$ positive sample}	
					\Else
						\State{$sample \leftarrow$ negative sample}
					\EndIf
					\State{Compute $\bm{t_i} - \bm{H_i\beta_{i-1}}[sample]$}
				\EndFor
			\EndFor
			\State{$\bm{\beta_i} \leftarrow \bm{\beta_{i-1}} + \bm{P_iH_i^T}(\bm{y_i} - \bm{H_i\beta_{i-1}})$}
		\EndFor
	\end{algorithmic}
\end{algorithm}
\end{figure}

In the skip-gram model, the desired graph embedding is obtained from weights of the neural network.
Specifically, we may be able to use the following weights for the graph embedding: 1) the input-side weights $\bm\alpha$, 2) the output-side weights $\bm\beta$, and 3) the average of $\bm\alpha$ and $\bm\beta$. 
Among them, the input-side weights are typically used for graph embedding.
However, since the input-side weights of the original OS-ELM are statically fixed at random values, in the proposed model we cannot directly use the input-side weights for the graph embedding. 
Although the original skip-gram model uses the input-side weights for the graph embedding, in the proposed model we utilize the trainable weights of OS-ELM (i.e., $\bm\beta$) to build the input-side weights as in~\cite{tied-weight}. 
Please note that although this technique is not suited for word2vec~\cite{tied-weight}, it can be applied for node2vec algorithm. Assume an activation function of the first layer is a linear function without bias vector. When we utilize $\bm\beta^T \in \mathbb{R}^{n \times N}$ as the input-side weights \footnote{In the skip-gram model we can assume $n=m$.}, the output probabilities are simply obtained by $O(\bm{x_i} \bm\beta^T \bm\beta)$, where $O$ is an activation function of the last layer such as sigmoid function. In this case, since $\bm{x_i}$ is one-hot vector where only a given center node is 1 and the others are 0, the output probability of the center node tends to be high. This is not suited for word2vec, because in the case of word2vec, probabilities that the center word appears as its neighboring words should be low; for example, when ``dog'' is a center word, ``dog'' rarely appears as neighboring words of the center word. In the case of node2vec, on the other hand, because of the nature of random walks described in Section~\ref{subsec:node2vec}, the same node often appears as its neighboring nodes.

In Figure~\ref{fig:models} (right), $\mu$ is a scale factor to transform $\bm\beta$ into the input-side weights. 
In this case, the input-side weights become a constant multiple of $\bm\beta$; thus the hidden-layer outputs $\bm{H_i}$ also become a constant multiple of the column vector corresponding to the center node which is extracted from $\bm\beta$. 
This eliminates the original random weights $\bm\alpha$ from OS-ELM, so we can reduce the model size and memory utilization.

The proposed model adopts the negative sampling~\cite{ns}. In this case, only a fraction of samples from negative nodes of teacher labels (i.e., nodes with a value of 0 in the one-hot vector) is trained instead of training all the negative samples. 
This can significantly reduce the training time by limiting the number of nodes to update, even if the number of nodes in the graph is huge. 
In general, 5 to 20 negative samples are sufficient for small datasets, while 2 to 5 negative samples are enough for large datasets~\cite{ns}. 
In Algorithm~\ref{alg:propose}, the innermost loop starting from line 9 corresponds to the negative sampling.
In this loop, $ns$ denotes the number of negative samples to be trained.
The outermost loop starting from line 1 processes $RW$ obtained from a random walk of node2vec. 
In the training phase, as described in Section~\ref{subsec:node2vec}, $RW$ is partitioned into samples (e.g., $N_S(u)$) by a given window size. 
In the case of $N_S(u)$, for example, node-$u$ is the center node, and nodes included in $N_S(u)$ are trained as positive nodes. 
Only a fraction of negative nodes is sampled randomly by the negative sampling method. 
The sampled frequency as negative nodes depends on the number of appearances of each node in the entire $RW$. 
This sampling is done by the Walker's alias~\cite{walker's}, which is a weighted sampling method. 
In this case, although the time complexity to build a table used in the sampling is proportional to the number of nodes, the time complexity of the sampling is $O(1)$.
In Algorithm~\ref{alg:propose}, lines 2 to 7 and lines 14 to 15 describe the training algorithm of OS-ELM.

\begin{figure}[t]
	\centering
	\includegraphics[keepaspectratio, scale=0.36]{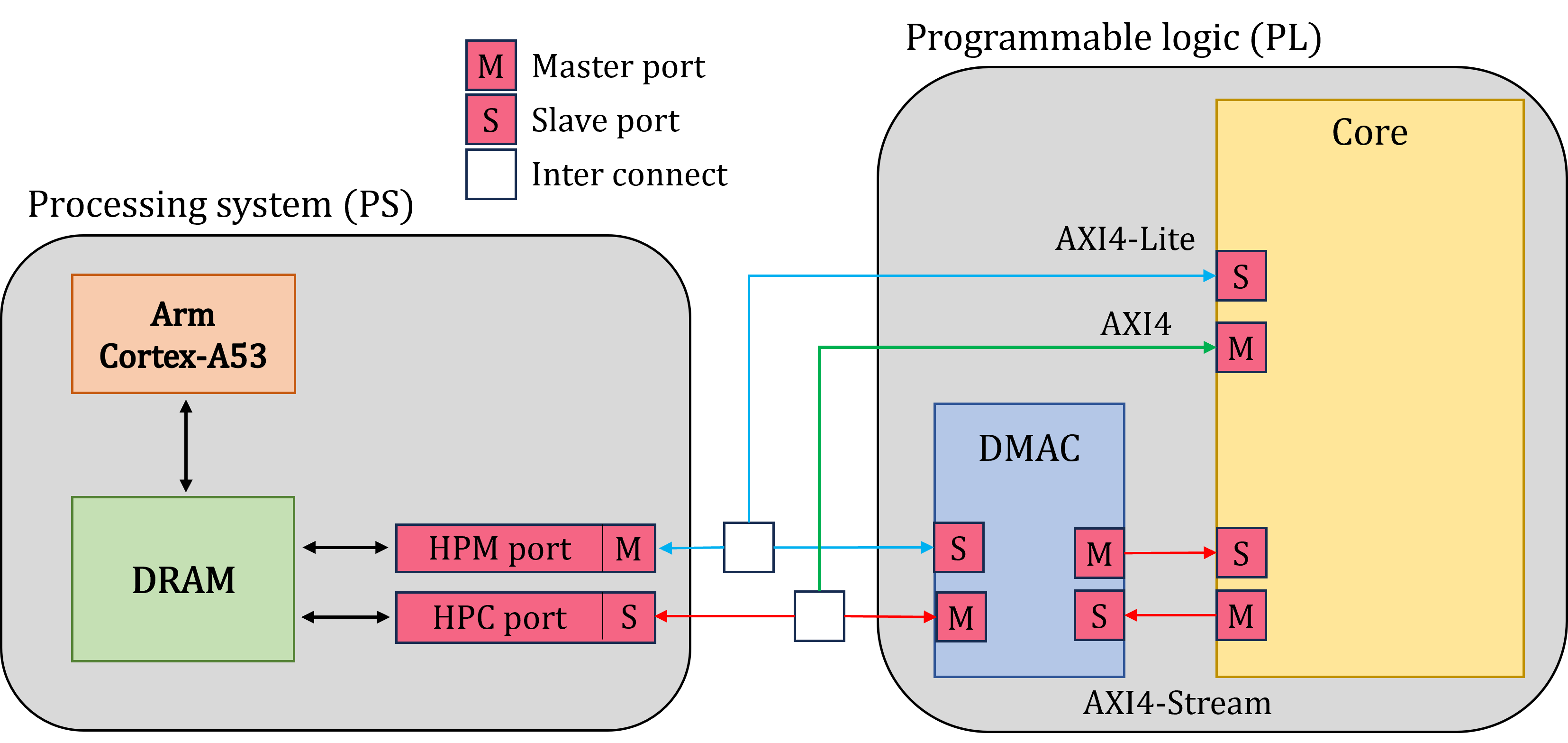}
	\caption{Board-level implementation}
	\label{fig:board_implementations} 
\end{figure}

\subsection{FPGA Implementation} \label{sec:fpga}
In this section, we describe an FPGA implementation of the proposed model.
We assume Xilinx Zynq MPSoC series as a target FPGA platform.
Figure~\ref{fig:board_implementations} illustrates a block diagram of the board-level implementation, which is divided into a processing system (PS) part and a programmable logic (PL) part.
Our sequentially-trainable node2vec accelerator is implemented in the PL part of the FPGA, which is denoted as ``Core'' in Figure~\ref{fig:board_implementations}.

\begin{figure}[t]
\begin{algorithm}[H]
	\caption{Modified algorithm for dataflow optimization}
	\label{alg:dataflow}
	\begin{algorithmic}[1]
		\For {each context}
			\State $Stage1:$
				\State $\quad \:\: \bm{H_i} \leftarrow \bm\beta[$center note$] \times \mu$
				\State $\quad \:\:$ Compute $\bm{P_{i-1}H_i^T}$ and $\bm{H_iP_{i-1}}$
			\State $Stage2:$
				\State $\quad \:\:$ Compute $\bm{P_{i-1}H_i^TH_iP_{i-1}}$ and $\bm{H_iP_{i-1}H_i^T}$
			\State $Stage3:$
			\Indent	
				\For {each window}
					\For {$itr=1$ to $ns+1$}
						\If {$itr = 1$}
							\State{$sample \leftarrow$ positive sample}	
						\Else
							\State{$sample \leftarrow$ negative sample}
						\EndIf
						\State{Compute $\bm{t_i} - \bm{H_i\beta_{i-1}}[sample]$}
					\EndFor
				\EndFor
			\EndIndent	
			\State $Stage4: $
				\State $\quad \:\: hpht\_inv \leftarrow \frac{1}{\bm{H_iP_{i-1}H_i^T}}$
				\State $\quad \:\: \bm{\Delta P} \leftarrow \bm{\Delta P} - \bm{P_{i-1}H_i^TH_iP_{i-1}} \times hpht\_inv$
				\State $\quad \:\: \bm{\Delta \beta} \leftarrow \bm{\Delta \beta} + \bm{P_iH_i^T}(\bm{y_i} - \bm{H_i\beta_{i-1}})$
		\EndFor
		\State $\bm{P_i} \leftarrow \bm{P_{i-1}} + \bm{\Delta P}$
		\State $\bm{\beta_i} \leftarrow \bm{\beta_{i-1}} + \bm{\Delta \beta}$
	\end{algorithmic}
\end{algorithm}
\end{figure}

As graphs become larger and the dimensions of graph embedding to be learned increase, it becomes challenging to implement all the weights on resource-limited FPGA devices. 
In the proposed model, since only a fraction of weights is updated by each training data by the negative sampling, only weights necessary for training are implemented on BRAM cells of the PL part. 
The training process is as follows. 
First, nodes are sampled from a graph using random walk by a host CPU in the PS part. 
The obtained result of a single random walk and negative samples necessary for training are pre-sampled by the CPU.
These samples are transferred from a DRAM to the BRAM via a DMA controller. 
After transferring the training data and negative samples, weights necessary for training (e.g., $\bm\beta$) are transferred from the DRAM to the BRAM. 
Then the model is sequentially trained in the PL part so that the weights are updated using these data. 
Finally, the trained weights are written back to the DRAM via the DMA controller. 
By repeating this procedure, a graph embedding can be trained. 
In our implementation, the same negative samples are used for multiple sets of training data as in~\cite{shared} to reduce the data transfer between DRAM and BRAM; in this case, training samples obtained by a single random walk are trained using the same negative samples.

To further speedup, a dataflow optimization is applied by modifying the update procedure of $\bm\beta$ in Algorithm~\ref{alg:propose}. 
Algorithm~\ref{alg:dataflow} shows the modified procedure. 
In Algorithm~\ref{alg:propose}, $\bm{P_i}$ and $\bm{\beta_i}$ are updated sequentially in each iteration of the outermost loop starting from line 1. 
Since there is a dependency between two successive iterations, a dataflow optimization cannot be applied in our original algorithm. 
In Algorithm~\ref{alg:dataflow}, on the other hand, $\bm P$ and $\bm\beta$ are updated outside the outermost loop (lines 19 and 20), and only their accumulated differences (i.e., $\bm{\Delta P}$ and $\bm{\Delta \beta}$) are updated sequentially inside the loop. 
This modification enables the dataflow optimization. 
Please note that the proposed model is trained with the same output-side weights $\bm\beta$ and the same intermediate data $\bm P$ for the result of a single random walk. 
It is expected that the proposed model can maintain an accuracy if the number of training data is sufficient, which will be evaluated in the next section.


\section{Evaluations} \label{sec:eval}
The proposed accelerator is implemented with Xilinx Vivado v2022.1 and Xilinx Vitis HLS v2022.1. 
We choose Xilinx Zynq UltraScale+ MPSoC series as a target FPGA platform; specifically, ZCU104 evaluation board (XCZU7EV-2FFVC1156) is used in this paper. 
As for software counterparts running on CPU, we use C/C++ to implement the models and compile them with gcc 9.4.0. In the performance evaluation, our FPGA implementation is compared with an embedded CPU of the FPGA board (ARM Cortex-A53 @1.2GHz) and a desktop computer (Intel Core i7-11700 @2.5GHz). Ubuntu 20.04.6 LTS in running on the computers. The clock frequency of the PL part of the FPGA board is set to 200MHz.

\begin{table}[t]
	\centering
	\caption{Three datasets used in evaluations}
	\label{table:dataset}
	\begin{tabular}{c|rrr} \hline
		Dataset                  & \multicolumn{1}{c}{\# nodes} & \multicolumn{1}{c}{\# edges} & \multicolumn{1}{c}{\# classes}\\ \hline
		Cora                          &  2,708                        &   5,429                       &    7\\ 
        Amazon Photo                  &  7,650                        & 143,663                       &    8\\ 
        Amazon Electronics Computers  & 13,752                        & 287,209                       &   10\\  \hline
	\end{tabular}
\end{table}

\subsection{Datasets} \label{subsec:dataset}
Table~\ref{table:dataset} lists three datasets used in our evaluations. 
We use Cora~\cite{cora}, Amazon Photo~\cite{amazoncp}, and Amazon Electronics Computers~\cite{amazoncp}. 
Cora is a paper citation network in a machine learning research field. 
Each node represents a paper, and each edge represents a citation relationship. 
Amazon Photo and Amazon Electronics Computers are subsets of Amazon co-purchase graph dataset~\cite{amazonproduct}.
Each node represents a product, and each edge represents that the two products are frequently bought together.

\begin{table}[t]
	\centering
	\caption{Hyper-parameters of node2vec}
	\label{table:node2vec_params}
	\begin{tabular}{c|cl}  \hline
		Parameter & Value & Description\\ \hline
		\textit{p} & 0.5 & Parameter to define $\alpha_{pq}(t, x)$\\
		\textit{q} & 1.0 & Parameter to define $\alpha_{pq}(t, x)$\\
		\textit{r} &  10 & Number of random walks per node\\
		\textit{l} &  80 & Length of single random walk\\
		\textit{w} &   8 & Window size\\
		\textit{ns} & 10 & Number of negative samples\\ \hline 
	\end{tabular}
\end{table}

\newcolumntype{C}{>{\centering\arraybackslash}p{10mm}}
\begin{table}[t]
	\centering
	\caption{Training time of a single random walk (vs. Cortex-A53 CPU)}
	\label{table:eval2_speed}
	\begin{tabular}{c|rrr} \hline 
		\multirow{2}{*}{}    	& \multicolumn{3}{c}{\# graph embedding dimensions}\\ \cline{2-4}
		                     	& \multicolumn{1}{C}{32} & \multicolumn{1}{C}{64} & \multicolumn{1}{C}{96}\\ \hline
		Original model on CPU (ms)           & 35.357 & 100.291 & 202.175\\ 
		Proposed model on CPU (ms)         & 18.753 & 35.941 & 72.612\\ 
		Proposed model on FPGA (ms)        & 0.777 & 0.878 & 0.985\\ \hline
		Speedup (vs. Original model on CPU)  & 45.504 & 114.227 & 205.254\\
		Speedup (vs. Proposed model on CPU) & 24.135 & 40.935 & 73.718\\ \hline
	\end{tabular}
\end{table}	

\newcolumntype{C}{>{\centering\arraybackslash}p{10mm}}
\begin{table}[t]
	\centering
	\caption{Training time of a single random walk (vs. Core i7 11700 CPU)}
	\label{table:eval2_speed_vs_i7}
	\begin{tabular}{c|CCC} \hline 
		\multirow{2}{*}{}       & \multicolumn{3}{c}{\# graph embedding dimensions}\\ \cline{2-4}
		                        & \multicolumn{1}{C}{32} & \multicolumn{1}{C}{64} & \multicolumn{1}{C}{96}\\ \hline
		Original model on CPU (ms)        &   1.309                &                 2.293  &                 3.285\\ 
		Proposed model on CPU (ms)      &   0.787                &                 1.426  &                 2.396\\ 
		Proposed model on FPGA (ms)        &   0.777                &                 0.878  &                 0.985\\ \hline
		Speedup (vs. Original model on CPU)  &   1.687                &                 2.612  &                 3.335\\
		Speedup (vs. Proposed model on CPU) &   1.013                &                 1.624  &                 2.432\\ \hline
	\end{tabular}
\end{table}	

\subsection{Execution Time} \label{sec:eval_speed}
Here, we evaluate the execution time of the proposed accelerator. 
The execution time is an elapsed time to train $RW$, which is obtained by a single random walk as mentioned in Section~\ref{sec:design}. 
In our evaluation, the length of random walk $l$ and the window size $w$ are set to 80 and 8, respectively. 
Thus, the training time of a single random walk is measured over 73 iterations of the outermost loop starting from line 1 in Algorithm~\ref{alg:dataflow}. 
Table~\ref{table:node2vec_params} summarizes the hyper-parameters of node2vec in this evaluation.
Table~\ref{table:eval2_speed} shows the execution times of the proposed accelerator and software implementations on ARM Cortex-A53 CPU. 
As shown, 1.89 to 2.77 times speedup is achieved by replacing the original skip-gram model with our OS-ELM based sequential model (Algorithm~\ref{alg:propose}). 
By implementing the proposed model on the FPGA, the proposed accelerator achieves 24.14 to 73.72 times speedup compared to that on ARM Cortex-A53 CPU. 
Compared to the CPU implementation of the original skip-gram model, our accelerator achieves 45.50 to 205.25 times speedup. 
In addition, Table~\ref{table:eval2_speed_vs_i7} shows the execution times of the proposed accelerator and software implementations on Intel Core i7 11700 CPU. 
Even when compared to the desktop computer, our small FPGA implementation achieves 1.01 to 3.34 times speedup.

\subsection{Accuracy} \label{sec:eval_accuracy}
For the accuracy evaluation, our trained graph embedding should be tested with a machine learning task. In this evaluation, it is used for a one-vs-rest logistic regression. 
The F1 score by the logistic regression is used as an evaluation metric. 
For the logistic regression, 90\% of the data are used as training data, and 10\% are used as test data for multiclass classification. 
SGD (Stochastic Gradient Descent) is used to train the original skip-gram model, and the learning rate is set to 0.01. 
In this evaluation, a graph embedding is trained three times. 
An average F1 score over the three trials is reported as the evaluation result.

\subsubsection{Impact of Dataflow Optimization} \label{sec:eval_fpga_accuracy}
\begin{figure}[t]
			\centering
			\includegraphics[keepaspectratio, scale=0.465]{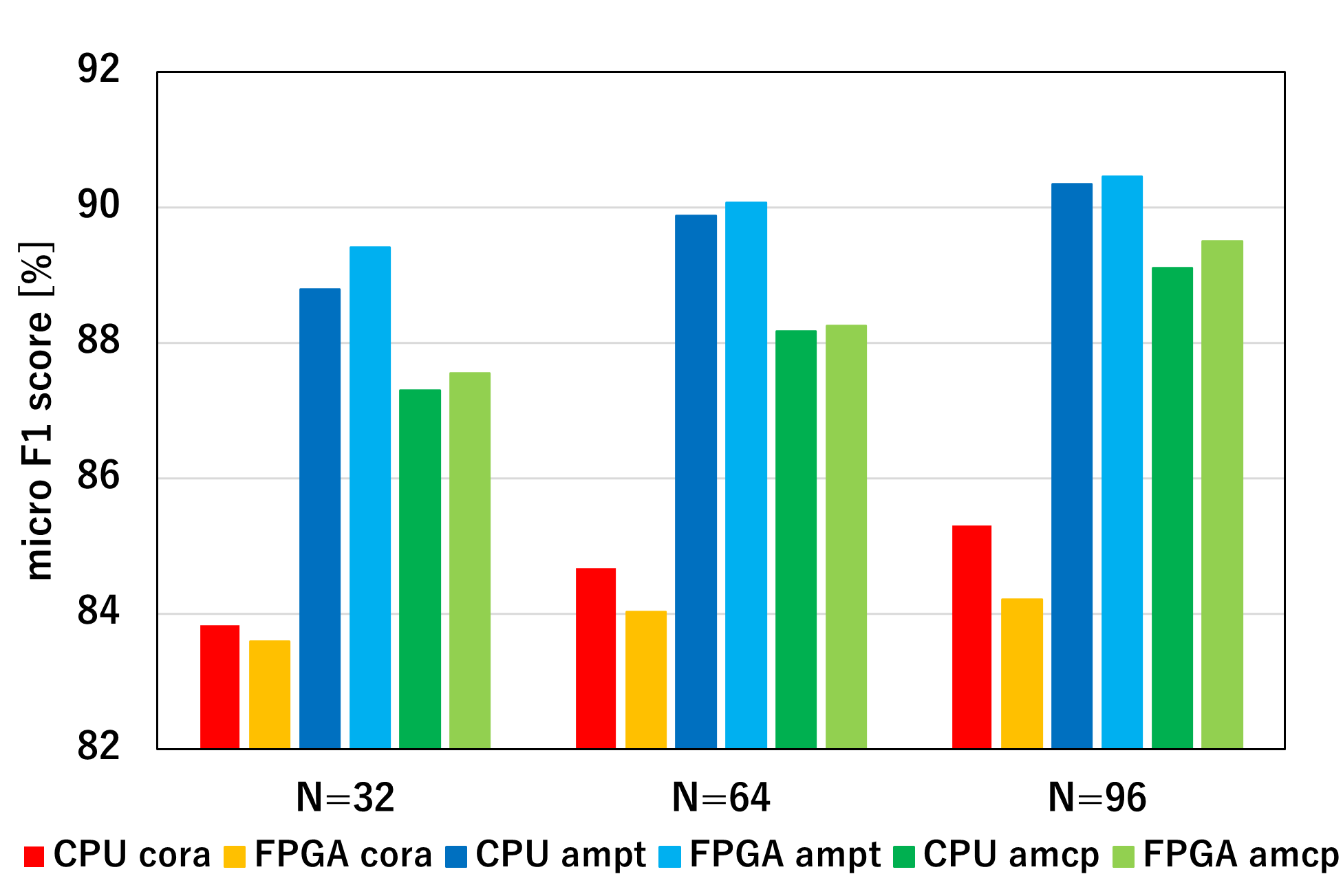}
			\label{fig:fpga_micro}
	\caption{Impact of dataflow optimization on accuracy}
	\label{fig:fpga_fscore} 
\end{figure}

To evaluate the impact of dataflow optimization applied to our FPGA accelerator, the proposed algorithm (Algorithm~\ref{alg:propose}) on CPU and the modified algorithm (Algorithm~\ref{alg:dataflow}) on FPGA are compared in terms of the accuracy. 
The three datasets described in Section~\ref{subsec:dataset} are used for this evaluation.
Figure~\ref{fig:fpga_fscore} shows the evaluation results, where ``ampt'' and ``amcp'' represent Amazon Photo and Amazon Electronics Computers datasets, respectively.
While the accuracy of the FPGA implementation decreases by up to 1.09\% in Cora dataset, no accuracy degradation is observed in the other two datasets, which have a relatively large number of nodes. 
In our FPGA implementation, the number of weight updates is decreased due to the dataflow optimization, and this affects the accuracy of Cora, which is a relatively small graph.

\begin{figure*}[h]
		\begin{minipage}{.333\textwidth}
			\centering
			\includegraphics[keepaspectratio, scale=0.38]{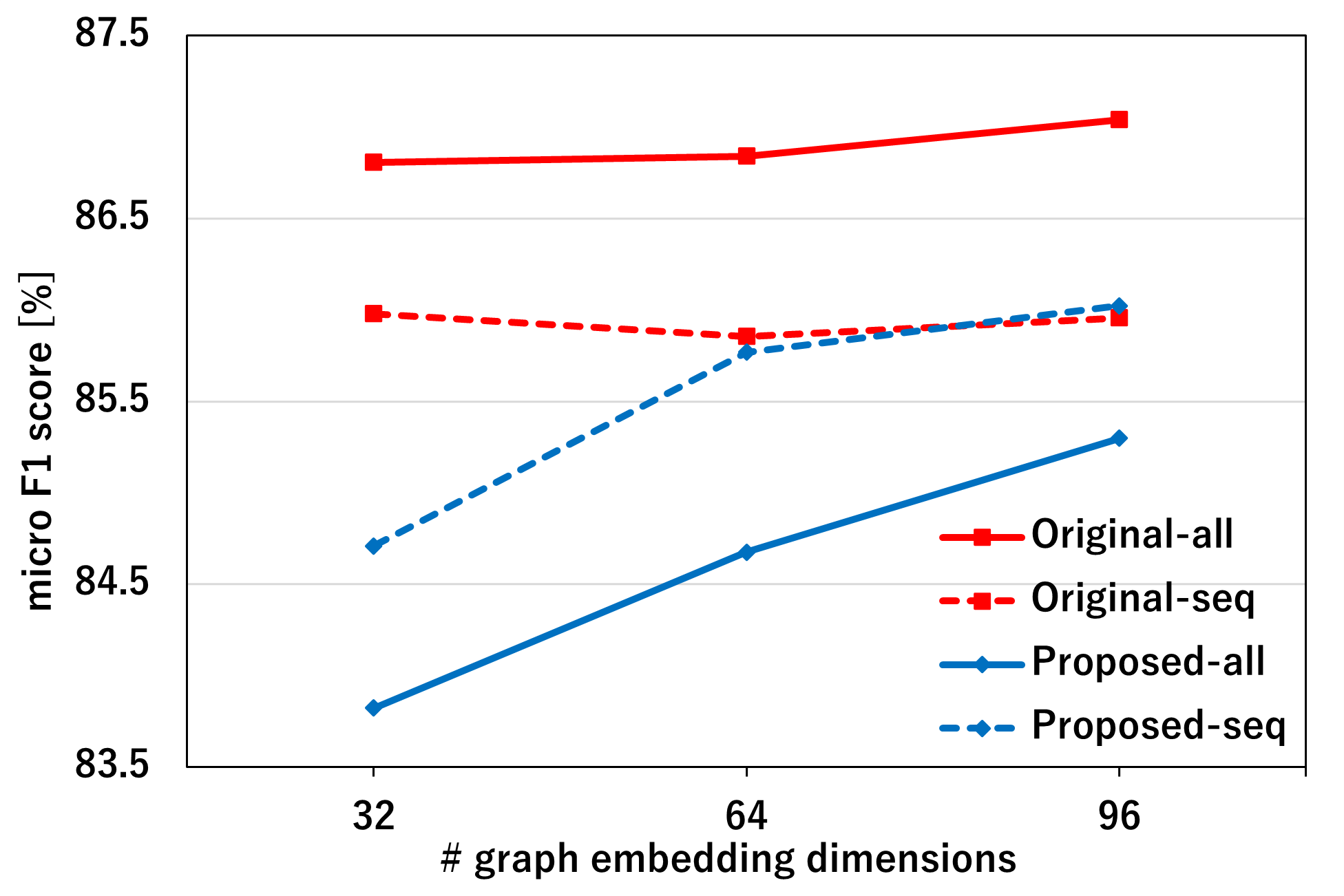}
			\subcaption{Cora }
			\label{fig:1}
		\end{minipage}
		\begin{minipage}{.333\textwidth}
			\centering
			\includegraphics[keepaspectratio, scale=0.38]{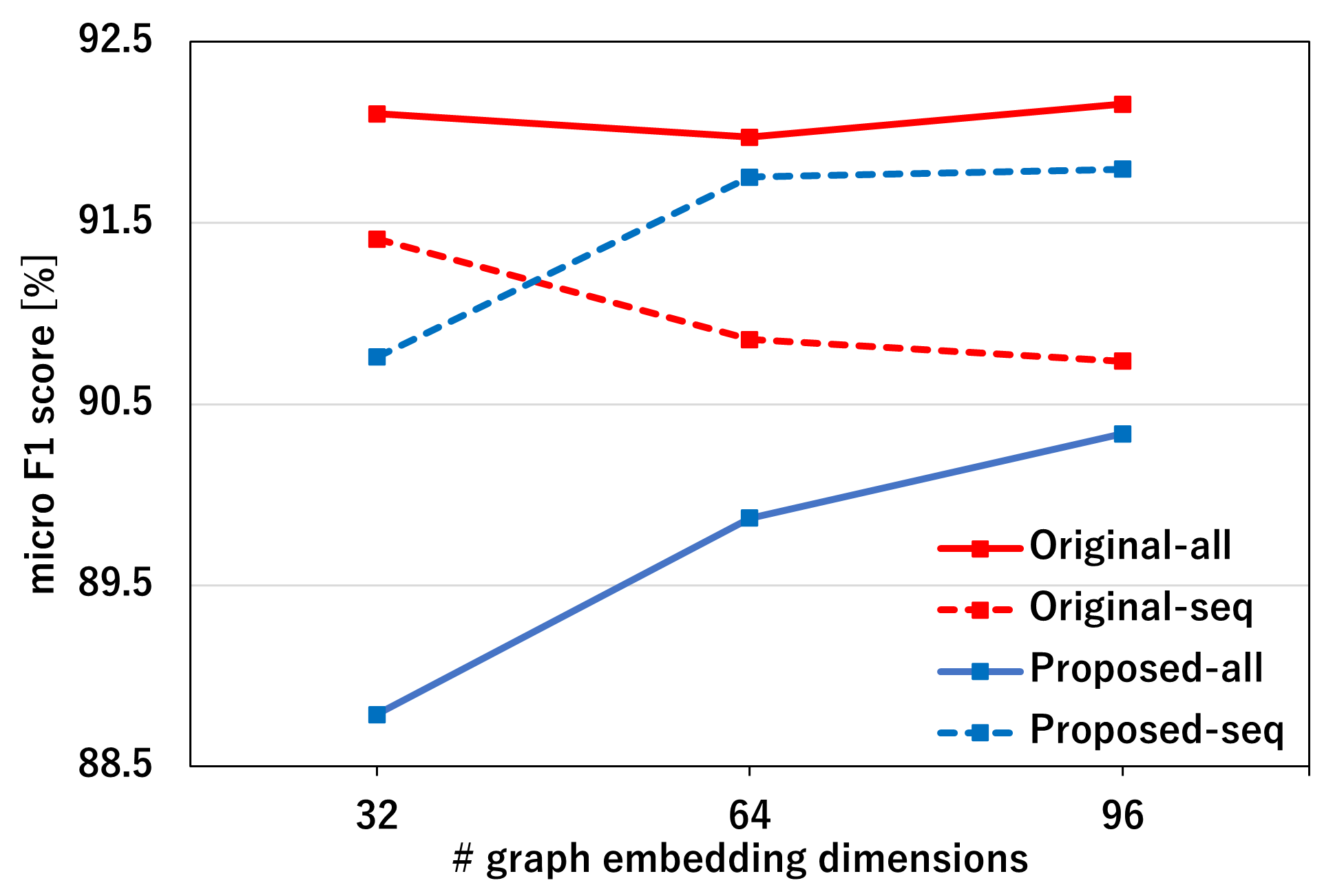}
			\subcaption{Amazon Photo }
			\label{fig:2}
		\end{minipage}
		\begin{minipage}{.333\textwidth}
			\centering
			\includegraphics[keepaspectratio, scale=0.38]{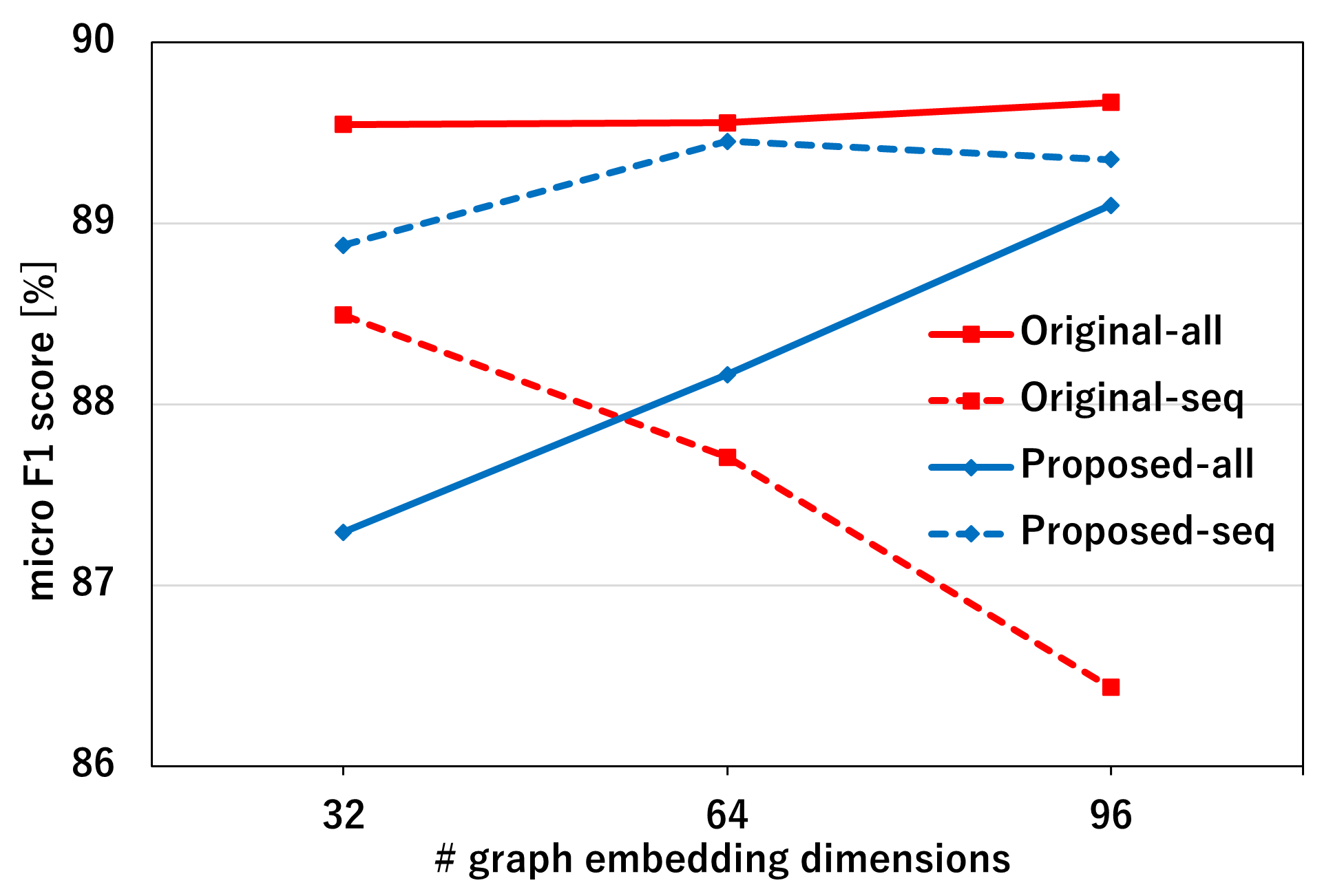}
			\subcaption{Amazon Electronics Computers }
			\label{fig:3} 
		\end{minipage} 
	\caption{Impact of sequential training on accuracy (micro F1 score)}
	\label{fig:seq_train} 
\end{figure*}

\begin{figure}[t]
	\centering
	\includegraphics[keepaspectratio, scale=0.455]{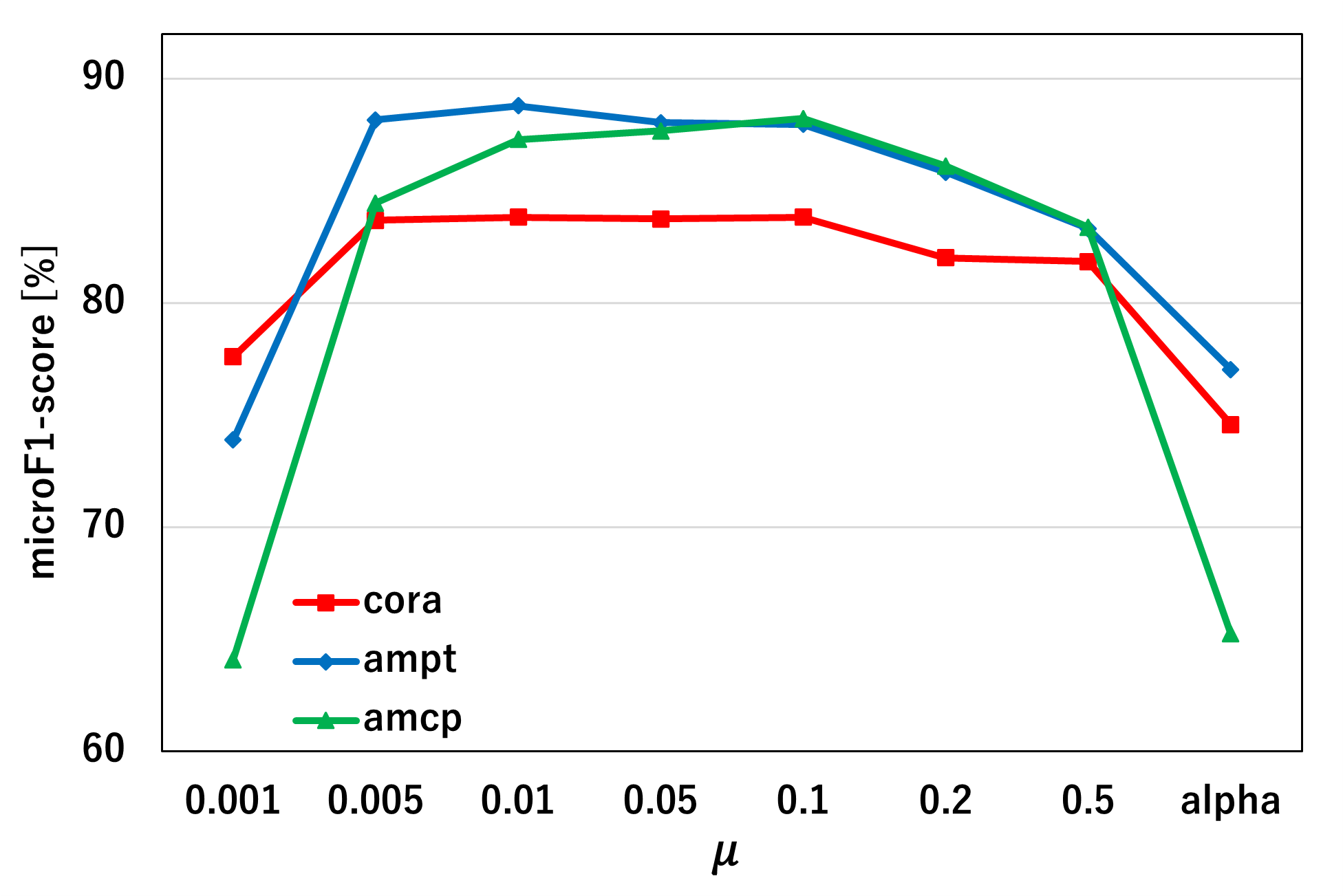}
	\caption{Impact on scale factor $\mu$ on accuracy}
	\label{fig:alpha_rate} 
\end{figure}

\subsubsection{Impact of Sequential Training} \label{sec:eval_seq_accuracy}
Next, we evaluate the benefit of the sequential training, which is one of major contributions of this paper. 
Figure~\ref{fig:seq_train} shows the evaluation results, where ``Original'' represents the original skip-gram model and ``Proposed'' represents our proposed model (i.e., Algorithm~\ref{alg:dataflow}). 
In addition, we examine two training scenarios: ``all'' and ``seq''. In the ``all'' case, an entire graph is trained assuming that all the edges exist from the beginning. 
In the ``seq'' case, only a fraction of edges is trained first; then, new edges are sequentially added to the graph, and a sequential training is executed every time a new edge is added. 
To build the initial graph of the ``seq'' case, we remove edges from an entire graph so that the initial graph becomes a forest without changing the number of connected components to the original entire graph. 
Subsequently, every time the removed edge is added, the random walk and training of node2vec are executed. 
In this case, the random walk starts from both the ends of an added edge.

As shown in Figure~\ref{fig:seq_train}, in the ``all'' case, the original skip-gram model achieves a higher accuracy compared with the proposed model for all the numbers of graph embedding dimensions (i.e., the numbers of hidden-layer nodes in the model) in all the datasets. 
In the ``seq'' case, on the other hand, the accuracy of our OS-ELM based sequentially-trainable model tends to be high compared to the original skip-gram model. 
In contrast, the accuracy of the original model drops when sequentially training the edges in the ``seq'' case. 
This implies that the sequential training using the backpropagation algorithm for the original model causes a catastrophic forgetting. 
This impact tends to be larger when the number of graph embedding dimensions increases and the graph becomes large. 
Although in this evaluation only a fraction of weights is updated by the negative sampling, the accuracy of the original model decreases due to the catastrophic forgetting. 
Please note that the proposed model in the ``seq'' case achieves a higher accuracy compared to the ``all'' case. 
Because in the ``seq'' case, a random walk and sequential training are executed every time a new edge is added, the number of training samples increases in the ``seq'' case; thus, the proposed sequential model successfully increases the accuracy. 
These results demonstrate that the graph embedding can be sequentially trained by using the proposed sequential model even if target graphs are large and dynamically updated.


\subsubsection{Impact of Scale Factor $\mu$} \label{sec:eval_alpha}
As proposed in Section~\ref{subsec:model}, in our sequential model, the input-side weights are replaced with a constant multiple of $\bm\beta$. 
Figure~\ref{fig:alpha_rate} shows the accuracy of the proposed model when the scale factor $\mu$ in Algorithm~\ref{alg:dataflow} is varied. 
Y-axis shows the accuracy, while X-axis shows the scale factor. 
The number of graph embedding dimensions is 32. 
In this graph, ``alpha'' represents an accuracy of a special case, where the input-side weights are fixed with random values as in the original OS-ELM algorithm. 
The accuracy of this ``alpha'' case is lower than our proposed model except when the scale factor $\mu$ is very small (i.e., 0.001). 
Actually, the accuracy of the proposed model when $\mu$ is 0.001 is quite low, indicating that a meaningful graph embedding may not be learned. 
On the other hand, we can see that the proposed model when $\mu \geq 0.005$ can learn a useful graph embedding. 
Especially, the accuracy is quite high when $\mu$ is ranging from 0.005 to 0.1, while it is gradually decreased when $\mu > 0.1$.


\subsection{Model Size} \label{sec:eval_memory}
Here, we compare the original skip-gram model and our proposed sequential model for FPGA in terms of the model size. 
Table~\ref{table:memory} shows their model sizes. 
The results show that the proposed model is up to 3.82 times smaller than the original model, thanks to our simplified OS-ELM based model, where the output-side weights $\bm\beta$ are reused for the input-side weights (thus we do not have to retain $\bm\alpha$). 
This reduces the memory consumption compared to the original skip-gram model; thus, our proposed model is beneficial for resource-limited IoT devices.

\subsection{FPGA Resource Utilization} \label{sec:eval_area}
Here, we evaluate the FPGA resource utilization of the proposed model. 
We use Zynq UltraScale+ XCZU7EV as a target FPGA device which has 11Mb BRAM and 1,728 DSP slices. 
Table~\ref{table:resource} shows the resource utilizations when the numbers of graph embedding dimensions are 32, 64, and 96, respectively.
In our FPGA implementation, the computational parallelism is basically set to 32. 
However, when the number of graph embedding dimensions is 64 and 96, the parallelism is partially set to 48 and 64 so that execution times of pipeline stages are equalized for the dataflow optimization. 
As shown in the table, when the number of graph embedding dimensions is 32, 79.80\% of DSP slices are consumed because fixed-point multiply-add operations are parallelized. 
When the number of graph embedding dimensions is 64, since the number of BRAM partitions is increased for further speedup, the BRAM and DSP utilizations are 86.86\% and 89.81\%, respectively. 

\begin{table}[t]
	\centering
	\caption{Model sizes of original model and proposed model (MB)}
	\label{table:memory}
	\begin{tabular}{c|c|rrr} \hline
		\# graph embedding  & \multirow{2}{*}{model} & \multirow{2}{*}{cora} &  \multirow{2}{*}{ampt}  & \multirow{2}{*}{amcp} \\
			 dimensions     &                        &                       &                         &                       \\ \hline
		\multirow{2}{*}{32} & Original model         &  1.350                &  3.823                  &  6.783                \\
                            & Proposed model         &  0.376                &  1.088                  &  1.897                \\ \hline
		\multirow{2}{*}{64} & Original model         &  2.676                &  7.559                  & 13.589                \\
                            & Proposed model         &  0.735                &  2.017                  &  3.600                \\ \hline
		\multirow{2}{*}{96} & Original model         &  3.999                & 11.295                  & 20.303                \\
                            & Proposed model         &  1.105                &  2.990                  &  5.318                \\ \hline
	\end{tabular}
\end{table}
\begin{table}[t]
	\centering
	\caption{Resource utilizations on XCZU7EV}
	\label{table:resource}
	\begin{tabular}{c|c|rrrr} \hline 
		\# graph embedding &  & \multicolumn{1}{c}{\multirow{2}{*}{BRAM}} & \multicolumn{1}{c}{\multirow{2}{*}{DSP}} & \multicolumn{1}{c}{\multirow{2}{*}{FF}} & \multicolumn{1}{c}{\multirow{2}{*}{LUT}} \\
           dimensions      &  &  &  &  & \\ \hline 
		\multirow{2}{*}{32} & Used    &    183         &        1,379    &       48,609    &     53,330     \\
							&  \%	 &  58.65         &        79.80    &        10.55    &      23.15     \\ \hline
		\multirow{2}{*}{64} & Used    &    271         &        1,552    &       77,584    &     87,901     \\
							&  \%	 &  86.86         &        89.81    &        16.84    &      38.15     \\ \hline
		\multirow{2}{*}{96} & Used    &    272         &        1,573    &       86,081    &    108,639     \\
							&  \%	 &  87.18         &        91.03    &        18.68    &      47.15     \\ \hline
	\end{tabular}
\end{table}


\section{Summary} \label{sec:conc}
In this paper, we proposed an OS-ELM based sequentially-trainable model for graph embedding and implemented it on an FPGA device.
Compared to the original skip-gram model, the proposed model achieved 1.89 to 2.77 times speedup. 
Furthermore, the FPGA implementation achieved 45.50 to 205.25 times speedup compared to the original model on ARM Cortex-A53 CPU. 
In the proposed model, by replacing the input-side weights with trained output-side weights (i.e., $\bm\beta$), we achieved both the accuracy improvement and the memory size reduction. 
For the sequential training of dynamic graphs, we showed that although the original model decreases the accuracy, the proposed model can be trained without decreasing the accuracy. 
In our future work, our FPGA-based sequentially-trainable model will be combined with an FPGA-based random walk implementation. We are also planning to compare our FPGA implementation with an embedded GPU implementation in terms of the execution time and energy efficiency in order to emphasize benefits of our FPGA-based sequential training approach.

{\bf Acknowledgments}
This work was partially supported by JST AIP Acceleration Research 
JPMJCR23U3, Japan.


\end{document}